\documentclass[10pt,twocolumn,letterpaper]{article}

\usepackage[pagenumbers]{cvpr} %

\usepackage{amsmath,amsfonts,bm}

\def\eqref#1{equation~\ref{#1}}

\def\1{\bm{1}}

\def\vs{{\bm{s}}}

\def\vz{{\bm{z}}}

\DeclareMathAlphabet{\mathsfit}{\encodingdefault}{\sfdefault}{m}{sl}
\SetMathAlphabet{\mathsfit}{bold}{\encodingdefault}{\sfdefault}{bx}{n}

\newcommand{\R}{\mathbb{R}}

\definecolor{cvprblue}{rgb}{0.21,0.49,0.74}
\usepackage[pagebackref,breaklinks,colorlinks,allcolors=cvprblue]{hyperref}

\usepackage[table]{xcolor}
\usepackage{multirow}
\usepackage{colortbl}
\usepackage{caption} %

\newcommand{\valpsnrcolor}[1]{%
  \ifdim #1 pt < 24pt \cellcolor{red!20}{#1}%
  \else\ifdim #1 pt < 24.5pt \cellcolor{orange!20}{#1}%
  \else\ifdim #1 pt < 25pt \cellcolor{yellow!20}{#1}%
  \else \cellcolor{green!20}{#1}%
  \fi\fi\fi
}

\newcommand{\testpsnrcolor}[1]{%
  \ifdim #1 pt < 21.5pt \cellcolor{red!20}{#1}%
  \else\ifdim #1 pt < 22pt \cellcolor{orange!20}{#1}%
  \else\ifdim #1 pt < 22.7pt \cellcolor{yellow!20}{#1}%
  \else \cellcolor{green!20}{#1}%
  \fi\fi\fi
}

\newcommand{\ssimcolor}[1]{%
  \ifdim #1 pt < 0.74pt \cellcolor{red!20}{#1}%
  \else\ifdim #1 pt < 0.75pt \cellcolor{orange!20}{#1}%
  \else\ifdim #1 pt < 0.76pt \cellcolor{yellow!20}{#1}%
  \else \cellcolor{green!20}{#1}%
  \fi\fi\fi
}

\newcommand{\lpipscolor}[1]{%
  \ifdim #1 pt > 0.20pt \cellcolor{red!20}{#1}%
  \else\ifdim #1 pt > 0.16pt \cellcolor{orange!20}{#1}%
  \else\ifdim #1 pt > 0.15pt \cellcolor{yellow!20}{#1}%
  \else \cellcolor{green!20}{#1}%
  \fi\fi\fi
}

\newcommand{\fidcolor}[1]{%
  \ifdim #1 pt > 90pt \cellcolor{red!20}{#1}%
  \else\ifdim #1 pt > 70pt \cellcolor{orange!20}{#1}%
  \else\ifdim #1 pt > 50pt \cellcolor{yellow!20}{#1}%
  \else \cellcolor{green!20}{#1}%
  \fi\fi\fi
}

\usepackage{float}
\usepackage{graphicx}

\usepackage[most]{tcolorbox}        %
\definecolor{violet}{HTML}{8E3BB8}
\definecolor{gray_}{HTML}{404040}
\definecolor{black_}{HTML}{000000}
\newtcolorbox[auto counter, number within=section]{summary}[1][]{
  colback=black_!5,
  colframe=gray_!80,
  colbacktitle=gray_!80,
  coltitle=white,
  fonttitle=\bfseries,
  title=Summary,
  arc=2pt,
  enhanced,
  #1
}
\newtcolorbox[auto counter, number within=section]{takeaway}[1][]{
  colback=black_!5,
  colframe=black_!80,
  colbacktitle=black_!80,
  coltitle=white,
  fonttitle=\bfseries,
  title=#1 Challenge,
  arc=2pt,
  enhanced,
}

\def\paperID{*****} %
\def\confName{CVPR}
\def\confYear{2025}

\title{Generative World Modelling for Humanoids \\[0.2em] \large 1X World Model Challenge Technical Report - Team Revontuli}
\author{
Riccardo Mereu$^{1,4}$\thanks{equal contribution.} \quad
Aidan Scannell$^{2*}$ \quad
Yuxin Hou$^{3}$ \quad
Yi Zhao$^{1}$ \\
Aditya Jitta$^{4}$ \quad
Antonio Dominguez$^{4}$ \quad
Luigi Acerbi$^{5}$ \quad
Amos Storkey$^{2}$ \quad
Paul Chang$^{4,5}$
\\[0.3em]
$^{1}$Aalto University \quad
$^{2}$University of Edinburgh \quad
$^{3}$Deep Render \quad
$^{4}$DataCrunch \quad
$^{5}$University of Helsinki
\\[0.4em]
\texttt{\small riccardo.mereu@aalto.fi, aidan.scannell@ed.ac.uk}
}

\begin{document}
\maketitle
\begin{abstract}
World models are a powerful paradigm in AI and robotics, enabling agents to reason about the future by predicting visual observations or compact latent states. 
The 1X World Model Challenge introduces an open-source benchmark of real-world humanoid interaction, with two complementary tracks: sampling, focused on forecasting future image frames, and compression, focused on predicting future discrete latent codes.
For the sampling track, we adapt the video generation foundation model Wan-2.2 TI2V-5B to video-state-conditioned future frame prediction. We condition the video generation on robot states using AdaLN-Zero, and further post-train the model using LoRA. For the compression track, we train a Spatio-Temporal Transformer model from scratch.
Our models achieve 23.0 dB PSNR in the sampling task and a Top-500 CE of 6.6386 in the compression task, securing 1st place in both challenges. 
\end{abstract}

\section{Introduction}
World models \citep{haRecurrentWorldModels2018} equip agents (\eg humanoid robots) with internal simulators of their environments.
By “imagining” the consequences of their actions, agents can plan, anticipate outcomes, and improve decision-making without direct real-world interaction.

A central challenge in world modelling is the design of architectures that are both sufficiently expressive and computationally tractable. Early approaches have largely relied on recurrent networks \citep{hafnerLearning2019,hafnerMasteringAtariDiscrete2022,hafnerMasteringDiverseControl2025} or multilayer perceptrons \citep{hansenTDMPC2ScalableRobust2023,hansenTemporalDifferenceLearning2022,scannell2025discrete,chuaDeepReinforcementLearning2018}. More recently, advances in generative modelling have driven a new wave of architectural choices. A prominent line of work leverages autoregressive transformers over discrete latent spaces \citep{bruceGenieGenerativeInteractive2024, robineTransformerbasedWorldModels2022, zhangSTORMEfficientStochastic2023, micheliTransformersAreSampleEfficient2022, barNavigationWorldModels2024, guoMineWorldRealTimeOpenSource2025}, while others explore diffusion- and flow-based approaches \citep{alonsoDiffusionWorldModeling2024, oasis2024}. At scale, these methods underpin powerful foundation models \cite{hong2022cogvideo, yang2024cogvideox, wan_paper, nvidiaCosmosWorldFoundation2025, HaCohen2024LTXVideo, kong2024hunyuanvideo} capable of producing realistic and accurate video predictions.

\begin{figure}[t]
\vskip -0.1in
    \centering
    \includegraphics[width=\linewidth]{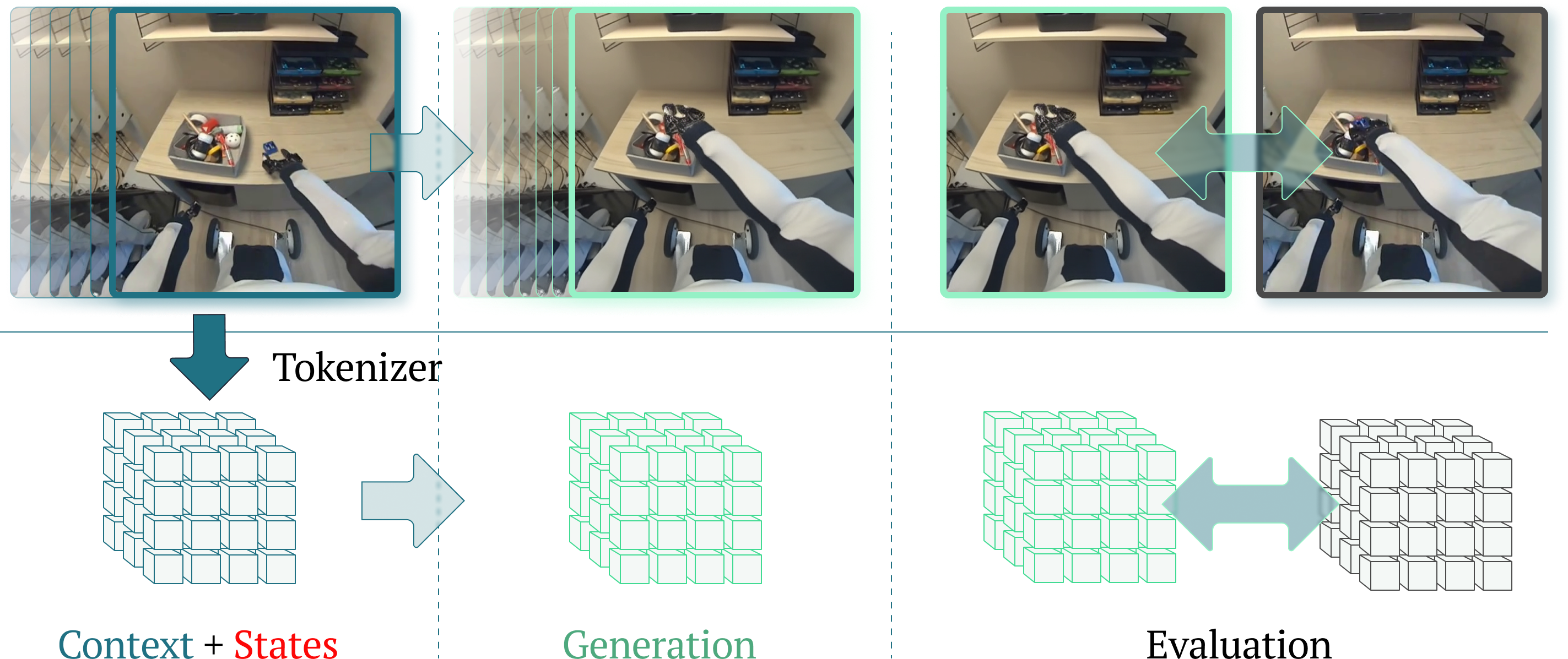}
    \caption{\textbf{Overview of the 1X World Model Challenges} 
Left depicts the context (inputs), middle the model generations, and right the evaluations.
\textbf{Sampling challenge (top)}: The model observes 17 past frames along with past and future robot states, then generates future frames in pixel space. Performance is measured by PSNR between the predicted and ground-truth 77th frame.
\textbf{Compression challenge (bottom)}: The Cosmos $8 \times 8 \times 8$ tokeniser encodes the history of 17 RGB frames into three latent token grids of shape $3 \times 32 \times 32$. Models must predict the next three latent token grids corresponding to the next 17 frames. Evaluation is based on Top-500 cross-entropy between predicted and ground-truth tokens.\looseness-1
}
    \label{fig:challenge_overview}
\end{figure}

\begin{table}[t]
\caption{\textbf{Performance on Public 1X World Model Leaderboard}}
\label{tab:leaderboard}
\centering
\scriptsize
\resizebox{\linewidth}{!}{ 
\begin{tabular}{l l cc cc c}
\toprule
\textbf{Benchmark} & \textbf{Submitter} 
& \multicolumn{2}{c}{\textbf{PSNR [dB]}} 
& \multicolumn{2}{c}{\textbf{CE loss}} 
& \textbf{Rank} \\
\cmidrule(lr){3-4} \cmidrule(lr){5-6}
 & & Test & Val & Test (Top-500) & Val & \\
\midrule
   & Revontuli & \textbf{23.00} & \textbf{25.53} & --     & --     & 1st \\
Sampling           & Duke      & 21.56          & 25.30             & --   & --     & 2nd \\
           & Michael   & 18.51          & --             & --     & --     & 3rd \\
\midrule
           & Revontuli & --             & --             & \textbf{6.64} & \textbf{4.92} & 1st \\
Compression & Duke      & --             & --             & 7.50   & 5.60     & 2nd \\
           & a27sridh  & --             & --             & 7.99   & --     & 3rd \\
\bottomrule
\end{tabular}
}
\vskip -0.1in
\end{table}

The \emph{1X World Model Challenge} evaluates predictive performance on two tracks: Sampling and Compression.
\cref{fig:challenge_overview} outlines the tasks, and \cref{tab:leaderboard} reports our results.
These challenges capture core problems when using world models in robotics.
Our methods show strong performance that we hope will shape future efforts.\looseness-1

\section{Sampling Challenge}
\label{sec:sampling}
\paragraph{Problem Statement}
In the sampling task, the model must predict the $512\times512$ frame observed by the robot $2$s into the future. Conditioning is provided by the first 17 frames  $\mathbf{x}_{0:16}$ and the complete sequence of robot states $\mathbf{s}_{0:76} \in \mathbb{R}^{77\times25}$. Performance is evaluated using PSNR between the predicted and ground-truth last frames.

\paragraph{Data Pre-processing}
We downsample the original 77 frames clips by a factor of four, yielding shorter 21 sample clips. As a result, this gives us five conditioning frames, ($\mathbf{x}_{0}, \mathbf{x}_{4}, \dots, \mathbf{x}_{16}$), and the remaining 16 serve as prediction targets. Wan2.2-VAE applies spatial compression to the first frame and temporal compression of 4 to the remaining frames, producing a latent sequence of length $(1+(L-1)/4)$ for a clip of length $L=21$.

\subsection{Model}
\paragraph{Base Model}
For our solution, we adapt Wan 2.2 TI2V-5B \cite{wan_paper}, a flow-matching generative video model with a 30-layer DiT backbone \cite{peebles2023dit}. 
The base model is designed as a text-image-to-video (TI2V), but we modified the architecture to condition the predictions on videos and robot states. The model operates on latent video representations from Wan2.2-VAE, which compresses clips to a size $(1+(L-1)/4)\times16\times16$.

\paragraph{Video-State Conditioning}
To incorporate video conditioning, we modified the masking of the input latents. In a standard image-to-video model, the first latent in the time dimension is masked, treating the input image as fixed during generation, thereby establishing a conditional mapping. We extend this idea by fixing multiple frames during generation, effectively transforming the model from image-to-video to video-to-video. The original Wan 2.2 also conditions textual prompts to generate videos. Since our dataset does not include textual descriptions, we use empty strings as text prompts while retaining the original cross-attention layer, enabling future work to leverage text conditioning.

\begin{figure}[t]
\vskip -0.2in
\centering
\includegraphics[width=0.9\linewidth]{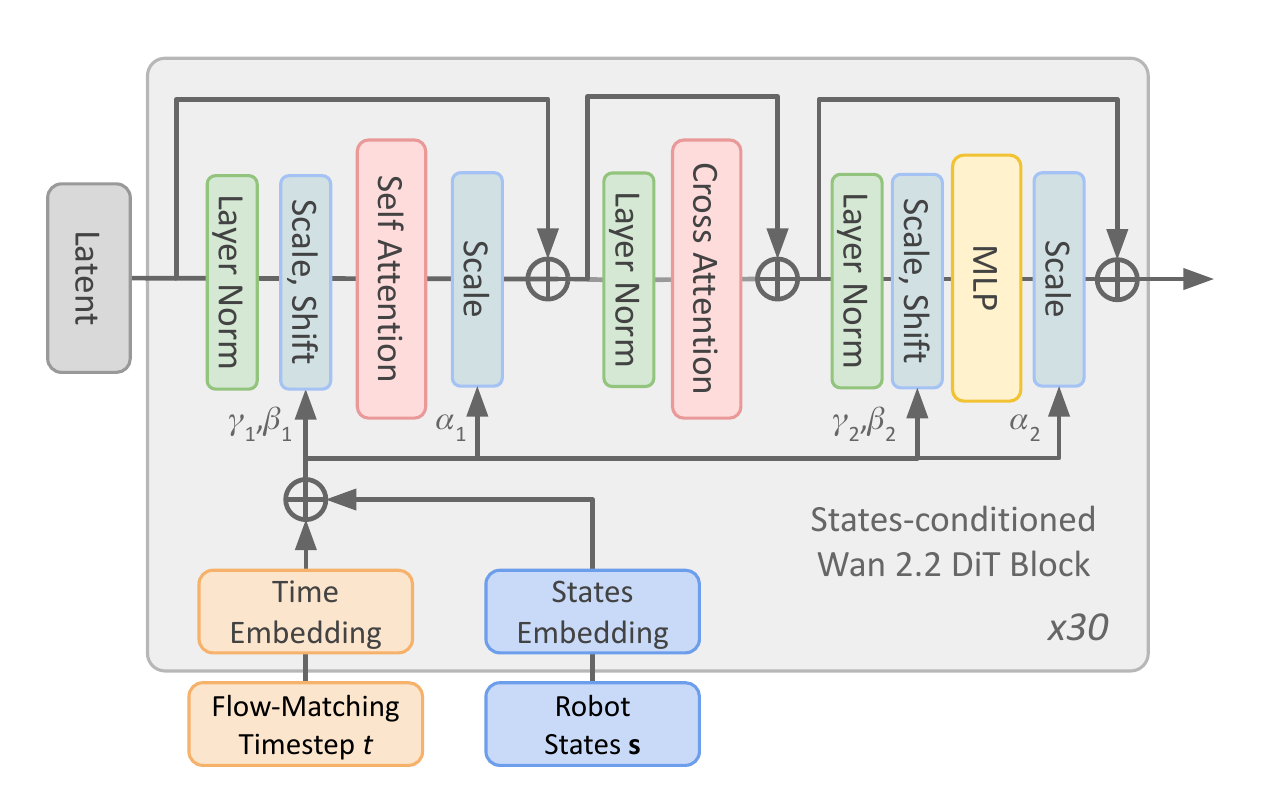}
\caption{\textbf{State conditioning of DiT-Block.} Wan2.2 TI2V-5B DiT architecture was updated to enable state conditioning using adaLN-Zero\cite{peebles2023dit} and combining it with the timestep of the Flow Matching scheduler \cite{wan_paper}.}
\label{fig:wan_arch}
\vskip -0.15in
\end{figure}

As shown in \cref{fig:wan_arch}, we incorporate state conditioning into the model’s predictions using adaLN-Zero~\cite{peebles2023dit} within Wan’s DiT blocks. We first downsample the states to match those of the downsampled video. The continuous angle and velocity states are augmented with sinusoidal features, and all states are projected through an MLP to a hidden dimension of $r_\text{dim}=256$.

Then, we compress the projected features along the temporal dimension with a 2-layer 1d convolutional network to match the compression of Wan-VAE for the video frames, mapping the state features to shape $((1+L//4), r_\text{dim})$. Finally, we fed the compressed feature into an MLP layer to get the modulation used by adaLN-Zero layers. The obtained robot modulation is added to the modulation of the flow matching timestep. The robot modulation acts differently on latent since the timestep embedding is the same for the whole latent, while for the states, they will modulate the latent slice associated with the corresponding frames.

\subsection{Training}
Models were trained for 23k steps with AdamW~\cite{loshchilov2018decoupled} with a constant learning rate of $4 \cdot 10^{-4}$. We applied LoRA~\cite{hu2022lora} fine-tuning with rank 32 on the Wan 2.2 DiT backbone. We experimented with and without classifier-free guidance (CFG)~\cite{ho2021classifierfree} during training but observed little improvement in PSNR performance (see \cref{sec:results}). Training was conducted on a DataCrunch instant cluster equipped with 4 nodes, each with $8 \times$ NVIDIA B200 GPUs. We used a total effective batch size of 1024. The B200 VRAM capacity of 184GB allows for more efficient training of memory-hungry video generation models. 

\subsection{Inference}
Since the challenge does not restrict inference compute time, we experimented with different approaches for our submissions. In our initial attempts, we followed \cite{liu2025duke} post-processing pipeline, applying Gaussian blur and performing histogram matching on the predicted frames. This post-processing improved the PSNR score by $1.2$dB, as reported in \cite{liu2025duke}. Because PSNR heavily penalizes outlier deviations from the target image, sharper images with slight errors are typically scored worse than blurrier images with comparable errors.\looseness-3

We found that exploiting predictive uncertainty with an ensemble of predictions outperformed Gaussian blurring. This produces blurring mainly in regions of high motion, such as the humanoid’s arms (see \cref{tab:sampling}). Increasing the number of ensemble samples improved PSNR on both the validation set and the public leaderboard, with different performance found from tuning the number of inference steps and the classifier-free guidance weight, as shown in \cref{tab:sampling}.

\begin{table}[t]
\vskip -0.1in
\centering
\caption{Sampling results on validation and test sets. $^\dagger$ The results on test set were obtained after the deadline. $^*$ This model has been trained on the whole train + validation raw dataset.}
\label{tab:sampling}
\resizebox{\linewidth}{!}{ 
\begin{sc}
\begin{tabular}{@{}ccccccccc@{}}
  \toprule
  Num. Inf.&Num.&CFG&Val.&Test&Val.&Val.&Val.\\
Samples & Samples &  Scale & PSNR~$[\uparrow]$& PSNR~$[\uparrow]$ & SSIM~$[\uparrow]$ & LPIPS~$[\downarrow]$ & FID~$[\downarrow]$ \\
  \midrule
  \multirow{4}{*}{20
  } 
   & 1  & --  & 22.63 & 21.05 & 0.707 & \textbf{0.137} & \textbf{40.23} \\
   & 5  & --  & 24.52 & 22.11 & 0.750 & 0.165 & 71.46 \\
   & 20 & --  & 24.88 & 22.42 & 0.762 & 0.201 & 90.71 \\
  \rowcolor{gray!15}
  \textbf{\textcolor{darkgray}{\textit{1st sub.}}}$^{*}$ 
   & 20 & -- & {\textcolor{darkgray}{26.62}} & {\textcolor{darkgray}{23.00}} & {\textcolor{darkgray}{0.836}} & {\textcolor{darkgray}{0.082}} & {\textcolor{darkgray}{31.70}} \\
  \midrule
  20  & 20 & 2.0 & 24.20 & 22.26 & 0.734 & 0.164 & 71.83 \\
  \rowcolor{gray!15}\textit{\textcolor{gray}{2nd sub.}} 
   & 20 & 1.5 & 24.59 & 22.53 & 0.746 & 0.169 & 74.10 \\
  \midrule
  \multirow{2}{*}{100} 
   & 20 & 1.5 & 25.07 & 22.55$^{\dagger}$ & 0.762 & 0.148 & 65.76 \\ 
   & 20 & 1.0 & \textbf{25.53} & \textbf{23.04}$^{\dagger}$ & \textbf{0.773} & 0.158 & 69.25 \\
  \bottomrule
\end{tabular}
\end{sc}
}

\vskip -0.1in
\end{table}

\subsection{Results}
\label{sec:results}
\cref{tab:sampling} reports the quantitative results of our model on the validation set using the PSNR metric. We further extend the evaluation by reporting Structural Similarity Index Measure (SSIM)~\cite{wang2003multiscale}, Learned Perceptual Image Patch Similarity (LPIPS)~\cite{zhang2018unreasonable}, and Fréchet Inception Distance (FID)~\cite{heusel2017fid}, all computed on our model’s predictions over the validation set.\looseness-1

The table is divided into three blocks. The first block contains models trained without classifier-free guidance (CFG)~\cite{ho2021classifierfree}. We ablate over the number of averaged samples used for final predictions, ranging from 1 to 20. Increasing the number of samples has a smoothing effect that improves \textsc{Val. PSNR} scores but degrades visual quality, as reflected in the other metrics. The bottom row of this block contains a model that is additionally trained on the validation dataset. This makes the values reported on the validation dataset not comparable with the rest of the entries in the table. However, the result on the public leaderboard showed a $+0.58$dB increase on PSNR.

The second and third blocks present models trained with CFG applied during training. Earlier experiments on the validation data showed that raising the cfg\_scale beyond a certain point did not improve PSNR scores. 
Nevertheless, we retained the run with cfg\_scale as our second-best competition submission.
For completeness, we also report results obtained by increasing the number of sampling steps using the same checkpoint. These results show consistent improvements over the previous CFG-based predictions.
\section{Compression Challenge}
\label{sec:compression}

\begin{figure*}[t] 
    \vskip -0.1in
    \centering
    \begin{subfigure}{0.7\linewidth}
        \centering
        \includegraphics[width=\linewidth]{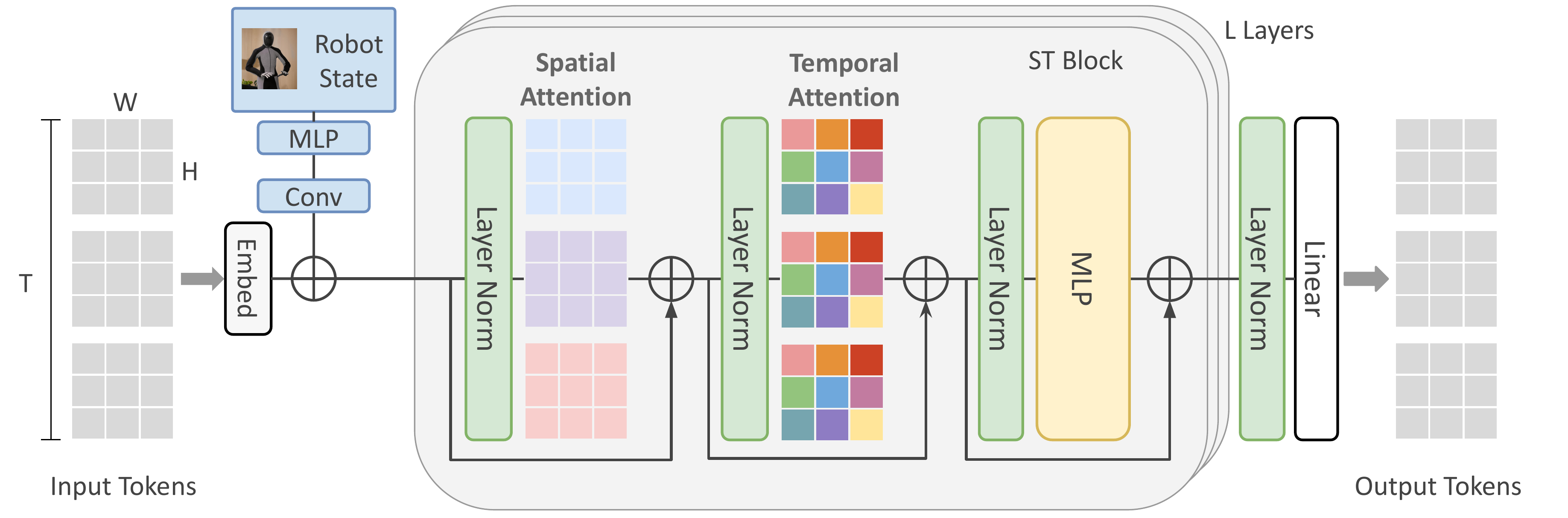}
        \caption{\textbf{Illustration of our ST-Transformer architecture for the compression challenge} Given three grids of past video tokens of shape $3 \times 32 \times 32$, as well as the robot state of shape $64 \times 25$ as context, the transformer predicts the future three grids of shape $3 \times 32 \times 32$. The ST-Transformer consists of $L$ layers of spatio-temporal blocks, each containing per time step spatial attention over the $H \times W$ tokens at time step $t$, followed by causal temporal attention of the same spatial coordinate across time, and then a feed-forward network. Each colour in the spatial and temporal attention represents a single self-attention map.}
    \label{fig:st-transformer-architecture}
    \end{subfigure}
    \hfill
    \begin{subfigure}{0.28\linewidth}
        \centering
        \includegraphics[width=\linewidth]{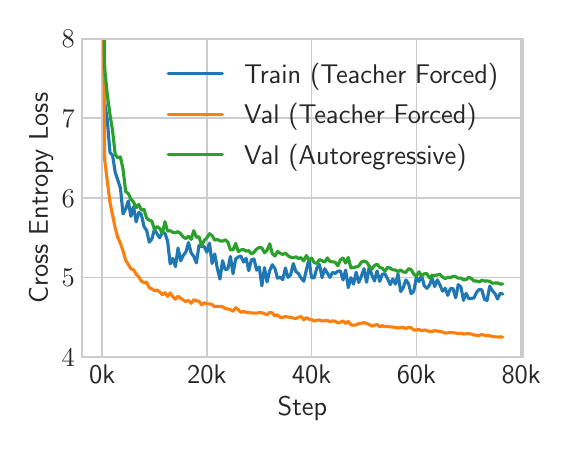}
        \caption{\textbf{Training curves for compression challenge} At train time, we use teacher forcing (blue). We then evaluate on the validation set using unrealistic teacher forcing (orange), as well as with the greedy autoregressive generation that will be used at inference time (green).}
        \label{fig:loss-curves-compression}
    \end{subfigure}
    \caption{Overall figure showing (a) the ST-Transformer world model architecture and (b) its training curves in the compression challenge.}
    \label{fig:combined}
    \vskip -0.1in
\end{figure*}

Unlike the Sampling Challenge, which measures prediction directly in pixel space, the Compression Challenge evaluates models in a discrete latent space. Each video sequence is first compressed into a grid of discrete tokens using the Cosmos $8 \times 8 \times 8$ tokeniser \citep{nvidiaCosmosWorldFoundation2025}, producing a compact sequence that can be modelled with sequence architectures.

\paragraph{Problem Statement}
Given a context of $H=3$ grids of $32 \times 32$ tokens and robot states for both past and future timesteps, the task is to predict the next $M=3$ grids of $32 \times 32$ tokens:
\begin{align}
\hat{\mathbf{z}}_{H:H+M-1} &\sim f_{\theta}(\vz_{0:H-1}, \vs_{0:63})
\end{align}
where $\hat{\vz}_{H:H+M-1}$ are the predicted token grids for the future frames.
The tokenized training dataset $\mathcal{D}$ contains approximately $306{,}000$ samples.
Each sample consists of:
\begin{itemize}
    \item \textbf{Tokenised video}: $6$ consecutive token grids (3 past, 3 future), each of size $32 \times 32$, giving $6144$ tokens per sample and $\sim1.88$B tokens overall.
    \item \textbf{Robot state}: a sequence $\vs \in \R^{64 \times 25}$ aligned with the corresponding raw video frames.
\end{itemize}
A block of three $32 \times 32$ token grids corresponds to 17 RGB frames at $256 \times 256$ resolution, so predictions in token space remain aligned with the original video. Performance is evaluated using top-500 cross-entropy, which considers only the top-500 logits per token.
\looseness=-2

\subsection{Model}
\paragraph{Spatio-temporal Transformer}
Following Genie \citep{bruceGenieGenerativeInteractive2024}, our world model builds on the Vision Transformer (ViT) \citep{dosovitskiyImageWorth16x162020,vaswaniAttentionAllYou2017}. An overview is shown in \cref{fig:combined}.
To reduce the quadratic memory cost of standard Transformers, we use a spatio-temporal (ST) Transformer \citep{xuSpatialTemporalTransformerNetworks2021}, which alternates spatial and temporal attention blocks followed by feed-forward layers. Spatial attention attends over $1 \times 32 \times 32$ tokens per frame, while temporal attention (with a causal mask) attends across $T \times 1 \times 1$ tokens over time. This design makes spatial attention, the main bottleneck, scale linearly with the number of frames, improving efficiency for video generation.
We apply pre-LayerNorm \citep{baLayerNormalization2016} and QKNorm \citep{henryQueryKeyNormalizationTransformers2020} for stability. Positional information is added via learnable absolute embeddings for both spatial and temporal tokens.
Our transformer used 24 layers, 8 heads, an embedding dimension of $512$, a sequence length of $T=5$, and dropout of $0.1$ on all attention, MLPs, and residual connections.

\paragraph{State Conditioning}
Robot states are encoded as additive embeddings following \citet{bruceGenieGenerativeInteractive2024}. The state vector is projected with an MLP, processed by a 1D convolution (kernel size 3, padding 1), and enriched with absolute position embeddings before being combined with video tokens.

\subsection{Training}
We implemented our model in PyTorch \citep{paszkePyTorchImperativeStyle2019} and trained it using the fused AdamW optimiser \citep{loshchilov2018decoupled} with $\beta_1=0.9$ and $\beta_2=0.95$ for $80$ epochs. Weight decay of $0.05$ was applied only to parameter matrices, while biases, normalisation parameters, gains, and positional embeddings were excluded.
Following GPT-2 \citep{radford2019language} and \citet{pressUsingOutputEmbedding2017,bertolottiTyingEmbeddingsYou2024}, we tied the input and output embeddings. This reduces the memory footprint by removing one of the two largest weight matrices and typically improves both training speed and final performance.\looseness-3

\paragraph{Training Objective}
The model was trained to minimise the cross-entropy loss between predicted and ground-truth tokens at future time steps:
\begin{align}
\min_\theta \, \mathbb{E}_{(\vz_t, \vs_t)_{t=0:K+M-1} \sim \mathcal{D}, \hat{\vz}_t \sim f_\theta(\cdot)} \left[ \sum_{t=K}^{K+M-1} \text{CE}\left( \hat{\vz}_t, \vz_t \right) \right], \nonumber
\end{align}
where $\hat{\vz}_t$ is the model output at time $t$, $\text{CE}$ denotes the cross-entropy loss over all tokens in the grid, and $\mathcal{D}$ is the dataset of tokenised video and state sequences.
Training used teacher forcing to allow parallel computation across timesteps, with a linear learning rate schedule from peak $8 \times 10^{-4}$ to $0$ after a warmup of $2000$ steps.

\paragraph{Implementation}
Training used automatic mixed precision (AMP) with \texttt{bfloat16}, but inference used \texttt{float32} due to degraded performance in \texttt{bfloat16}. Linear layer biases were zero-initialised, and weights (including embeddings) were drawn from $\mathcal{N}(0,0.02)$. We trained with an effective batch size of 160 on the same B200 DataCrunch instant cluster as in the sampling challenge.\looseness-2

\subsection{Inference}
Our autoregressive model generates sequences via
\begin{align}
p(\vz_{H:H+M-1} \mid \vz_{0:H-1}, \vs_{0:63}) = \prod_{t=H}^{H+M-1} f_{\theta}(\vz_{t} \mid \vz_{<t}, \vs_{0:63}), \nonumber
\end{align}
where each step outputs a categorical distribution over each spatial token.
\emph{Sampling} draws $\vz_t \sim f_{\theta}(\cdot)$, introducing diversity but typically yields lower-probability trajectories and higher loss.  
\emph{Greedy decoding} instead selects 
\[
\vz_t = \arg\max_{\vz} f_{\theta}(\vz \mid \vz_{<t}, \vs_{0:63}),
\]
producing deterministic, high-probability sequences that we found both effective and efficient.  

\subsection{Results}
\cref{fig:loss-curves-compression} shows the training curves for our ST-Transformer. The blue curve corresponds to the training loss under teacher-forced training. While the teacher-forced validation loss is optimistic -- since it conditions on ground-truth inputs -- it can be interpreted as a lower bound on the achievable loss, representing the performance of an idealised autoregressive model with perfect inference.
To reduce the gap between teacher-forced and autoregressive performance, we experimented with scheduled sampling \citep{bengioScheduledSamplingSequence2015a,mihaylovaScheduledSamplingTransformers2019}. However, this did not lead to meaningful improvements.\looseness-3

\section{Conclusion}
In this report, we presented two complementary approaches that achieved strong performance across both 1X World Model Challenges.
First, we showed how internet-scale data can be leveraged by fine-tuning a pre-trained image–text-to-video foundation model. Using multi-node training on the DataCrunch instant cluster, we reached first place on the leaderboard in only ~36 hours—an order of magnitude faster than the runner-up, who required about a month. To further improve inference, we averaged over samples to selectively blur regions of high predictive uncertainty. While this proved effective for optimising PSNR, the most suitable inference strategy for downstream decision-making remains an open question.
Second, we demonstrated how a spatio-temporal transformer world model can be trained on the tokenised dataset in under 17 hours. We found that greedy autoregressive inference offered a practical balance of speed and accuracy.
Despite its simplicity, the model achieved substantially lower loss values than other leaderboard entries.

\clearpage
{
    \small
    \bibliographystyle{ieeenat_fullname}

}
\end{document}